\documentclass[lettersize,journal]{IEEEtran}
\usepackage{amsmath,amsfonts}
\usepackage{algorithm}
\usepackage{algpseudocode}
\usepackage{array}
\usepackage[caption=false,font=normalsize,labelfont=sf,textfont=sf]{subfig}
\usepackage{textcomp}
\usepackage{stfloats}
\usepackage{url}
\usepackage{verbatim}
\usepackage{graphicx}
\usepackage{cite}
\usepackage{booktabs}
\usepackage{pifont}
\usepackage{multirow}
\usepackage{makecell}
\usepackage{colortbl}
\usepackage{arydshln}
\usepackage[backref]{hyperref} 
\hypersetup{hidelinks}

\begin{document}

\title{FMReward: Aligning and Evaluating Audio-Driven 3D Facial Animation with Human Preferences}

\author{Sijing Wu, Yunhao Li, Zhilin Gao, Huiyu Duan, Yucheng Zhu, \\ Guangtao Zhai,~\IEEEmembership{Fellow,~IEEE}, and Patrick Le Callet,~\IEEEmembership{Fellow,~IEEE}
\thanks{Sijing Wu, Yunhao Li, Zhilin Gao, Huiyu Duan, and Guangtao Zhai are with the Institute of Image Communication and Network Engineering, Shanghai Jiao Tong University, Shanghai, China (e-mail: \{wusijing, lyhsjtu, undefined49527, huiyuduan, zhaiguangtao\}@sjtu.edu.cn).}
\thanks{Yucheng Zhu is with the USC-SJTU Institute of Cultural and Creative Industry, Shanghai Jiao Tong University, Shanghai, China (e-mail: zyc420@sjtu.edu.cn).}
\thanks{Patrick Le Callet is with the Institut Universitaire de France (IUF), University of Nantes, France (e-mail: patrick.lecallet@univ-nantes.fr).}
}


\maketitle

\begin{abstract}
Audio-driven 3D facial animation is essential for advancing immersion and interactivity in virtual experiences. 
Although recent advances have shown promising capabilities, the training and evaluation of existing methods typically rely on ground-truth-based errors, which fall short of aligning with human preferences.
To address this, we present a comprehensive framework that learns an automatic perceptual model from human preference data and leverages it to improve and evaluate the perceptual quality of audio-driven 3D facial animation.
To begin with, we construct \textbf{FMPair} (\underline{\textbf{F}}acial \underline{\textbf{M}}otion \underline{\textbf{Pair}}wise preference), the first human preference dataset for audio-driven 3D facial animation, which is built through a systematic annotation pipeline and comprises 65,574 annotated 3D facial motion pairs from 8,834 distinct in-the-wild audio clips.
Based on the pairwise comparison dataset, we propose a \underline{\textbf{F}}acial \underline{\textbf{M}}otion \underline{\textbf{Reward}} model, termed \textbf{FMReward}, which takes audio and 3D facial motion as inputs and predicts a perceptual quality score aligned with human preferences. Building upon FMReward, we further introduce \underline{\textbf{F}}acial \underline{\textbf{M}}otion reward \underline{\textbf{F}}eedback \underline{\textbf{L}}earning (\textbf{FMFL}), a direct fine-tuning algorithm that leverages a pretrained reward model to optimize diffusion-based audio-driven 3D facial animation models for better alignment with human preferences.
Extensive experiments demonstrate the superiority of FMReward over other metrics in aligning with human preferences and the effectiveness of FMFL in improving the perceptual quality of audio-driven 3D facial animation.
The dataset and codes will be released at: \url{https://github.com/wsj-sjtu/FMReward}.

\end{abstract}


\section{Introduction}
\begin{figure}[t]
\centering
\includegraphics[width=\linewidth]{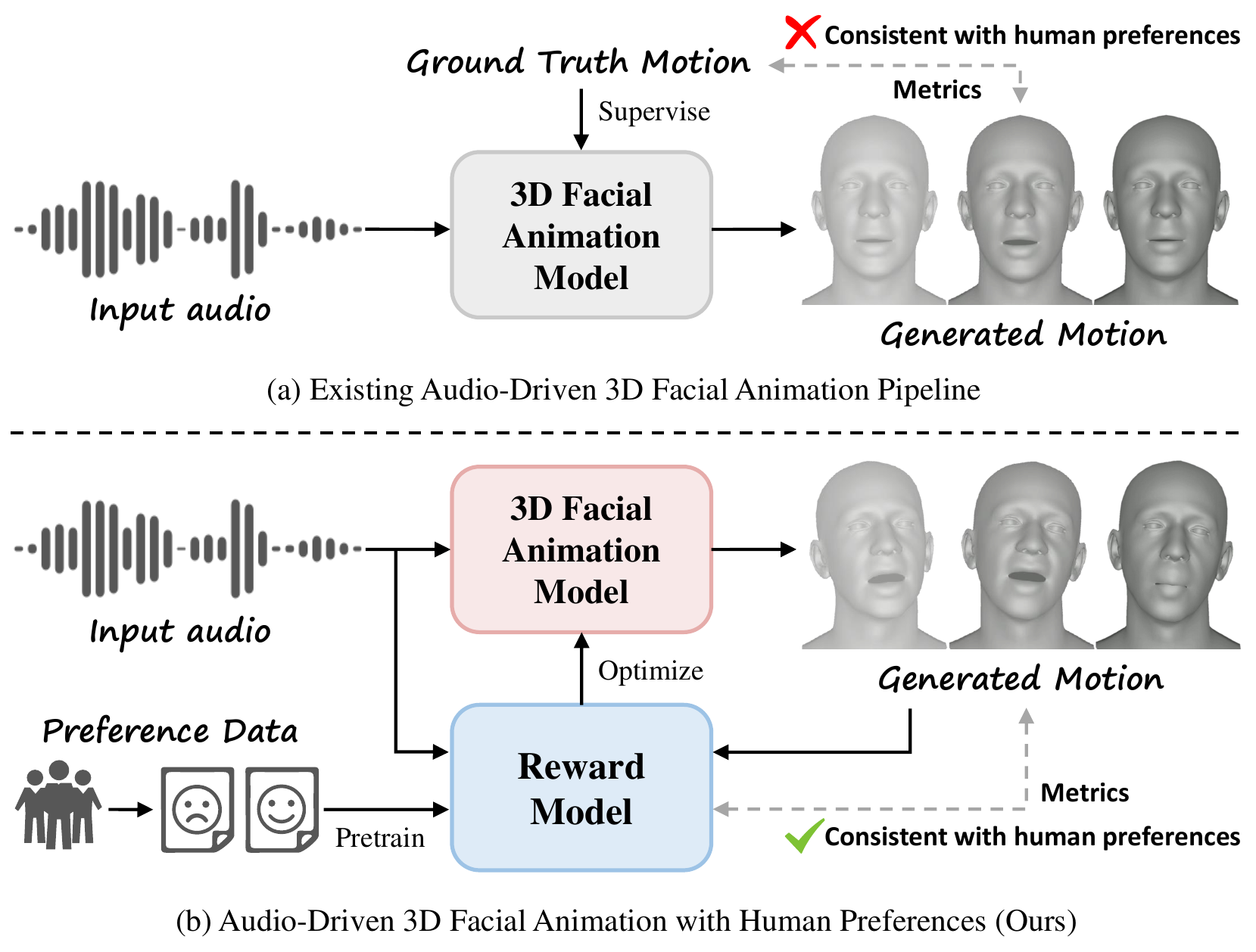}
\caption{Comparison of existing audio-driven 3D facial animation pipeline (a) and our human preference-based pipeline (b). We first train a reward model using pairwise human preference data, which is then used to fine-tune the pretrained audio-driven 3D facial animation model for better alignment with human preferences.}
\vspace{-2mm}
\label{fig:teaser}
\end{figure}

\IEEEPARstart{A}{udio-driven} 3D facial animation has attracted significant attention in computer graphics and computer vision owing to its critical role in advancing realism and interactive experiences across a wide range of applications such as AR/VR content creation, telepresence, and games.
Recent advances \cite{peng2023selftalk,sun2024diffposetalk,chu2025artalk,liu2024emage,zhang2025speechact} have demonstrated remarkable progress in producing lip-synchronized facial motion from speech audio within the ground-truth-based paradigm.
However, synchronized lip movements alone are far from satisfactory for practical applications \cite{xing2023codetalker,sun2024diffposetalk,wu2024mmhead}, as naturalness, expressiveness, and artifacts also substantially impact user experience. These perceptual dimensions are difficult to quantify using conventional objective metrics owing to their subjectivity and the nonlinear nature of facial dynamics.
These limitations highlight the significance of incorporating human preference to advance audio-driven 3D facial animation toward better alignment with human perception.

Existing works typically rely on ground-truth supervision and simple heuristic criteria for both training and evaluation.
In training, mean squared error (MSE) between generated and ground-truth 3D facial motion parameters or vertex sequences is commonly employed as the objective function \cite{cudeiro2019capture,fan2022faceformer,xing2023codetalker,peng2023selftalk,thambiraja2023imitator,wu2024mmhead}. Various auxiliary objectives, such as velocity matching to the ground truth \cite{cudeiro2019capture,peng2023selftalk,thambiraja2023imitator,wu2025singinghead,sun2024diffposetalk} and smoothness regularization of the generated facial motion \cite{sun2024diffposetalk}, are also incorporated to improve temporal consistency. However, these objectives mainly capture low-level geometric errors or synchronization, which inherently fail to reflect human perception, often leading to stiff and unnatural results.
In evaluation, the most commonly used metrics include ground-truth-based measures such as lip vertex error (LVE) \cite{richard2021meshtalk} and upper-face dynamics deviation (FDD) \cite{xing2023codetalker}, as well as heuristic measures like beat align score (BA) \cite{siyao2022bailando,sun2024diffposetalk} and diversity \cite{wu2024mmhead,sun2024diffposetalk}. However, these metrics fail to align with human perception due to the many-to-many mapping between audio and facial motion, as well as the inherent limitations of heuristic definitions.
Although user studies have been employed \cite{fan2022faceformer,xing2023codetalker,li2024pose,song2024expressive,sun2024diffposetalk,gu2025diverse,xuan2023narrator}, they are costly, difficult to generalize across scenarios, and cannot directly guide training.
The aforementioned limitations underscore the necessity of automatic perceptual modeling to capture human preference and guide both training and evaluation.

To fill this gap, we propose the first human preference dataset for audio-driven 3D facial animation and an automatic perceptual model trained on it to approximate human evaluation.
Specifically, we first construct the \textbf{FMPiar} dataset, which consists of 65,574 annotated 3D facial motion pairs from 8,834 distinct audio clips across diverse languages. The audio samples are drawn from the validation and test splits of two in-the-wild audio-driven 3D facial animation datasets, TFHP \cite{sun2024diffposetalk} and MMHead \cite{wu2024mmhead}. The 3D facial motion pairs are derived from 101,216 unique sequences, including ground-truth facial motions and diverse generations from four state-of-the-art methods \cite{sun2024diffposetalk,fan2024unitalker,wu2024probtalk3d,wu2024mmhead}, to ensure the diversity of the dataset. 
Based on the FMPair dataset, we propose a perceptual model, \textbf{FMReward}, which takes audio and 3D facial motion as inputs and predicts a perceptual quality score aligned with human preferences. Specifically, FMReward adopts a simple and efficient transformer-based framework, which fuses audio and motion features through cross-attention and subsequently regresses a quality score from the fused representation.
The training objective is based on the Bradley–Terry model \cite{bradley1952rank,wang2024aligning}, leveraging pairwise preference data as probabilistic supervision to align the model's judgments with human evaluation.
Comprehensive experiments demonstrate that FMReward surpasses widely used objective metrics and quality assessment methods in aligning with human preferences.

Building upon FMReward, we further introduce \textbf{FMFL}, a direct fine-tuning algorithm to optimize diffusion-based audio-driven 3D facial animation models for better alignment with human preferences.
Concretely, we utilize the pretrained reward model to evaluate the sampling results and compute losses, which are used to optimize the diffusion model during fine-tuning.
To reduce computational cost and improve stability, denoising starts from a randomly selected later step, and gradients are computed only at the final step. Moreover, at each fine-tuning step, the original training loss and the reward loss are back-propagated alternately to mitigate overfitting.
We conduct experiments on our self-constructed vanilla diffusion model based on DiffPoseTalk \cite{sun2024diffposetalk}, referred to as \underline{T}alking \underline{D}iffusion \underline{M}odel (TDM).
Both quantitative and qualitative evaluations demonstrate the effectiveness of FMFL in improving diffusion-based audio-driven 3D facial animation models toward better alignment with human preferences.

In summary, the main contributions of this paper are:
\begin{itemize}
    \item We present the first human preference dataset for audio-driven 3D facial animation, \textbf{FMPair}, which contains 65,574 annotated 3D facial motion pairs from 8,834 distinct audio clips.
    \item We propose \textbf{FMReward}, the first automatic perceptual model for audio-driven 3D facial animation that predicts quality scores aligned with human preferences.
    \item We introduce \textbf{FMFL}, a direct fine-tuning algorithm that optimizes diffusion-based audio-driven 3D facial animation models to generate facial motions with better perceptual quality.
    \item Extensive experiments demonstrate the superiority of FMReward in aligning with human preferences and the effectiveness of FMFL in improving the perceptual quality of audio-driven 3D facial animation.
\end{itemize}

\section{Related Work}
\subsection{Audio-driven 3D Facial Animation}
Audio-driven 3D facial animation aims to generate synchronized and natural 3D facial motion sequences from input speech audio. In recent years, it has attracted increasing attention due to its numerous applications in AR/VR content creation and virtual avatar animation \cite{wu2023ganhead, qian2024gaussianavatars}.
Audio-driven 3D facial animation methods can be broadly categorized into rule-based methods \cite{taylor2012dynamic,xu2013practical,edwards2016jali} and learning-based methods \cite{cudeiro2019capture,richard2021meshtalk,zhang20213d,liu2021geometry,fan2022faceformer,chai2022personalized,xing2023codetalker,peng2023selftalk,wu2023speech,stan2023facediffuser,aneja2024facetalk,yang2024probabilistic,sun2024diffposetalk,ma2024diffspeaker,pan2024expressive,song2024expressive,li2024pose,zhuang2024learn2talk,song2024talkingstyle,pan2025vasa,gu2025diverse}.
Rule-based methods generally define intricate correspondences between phonemes and lip movements. While these methods achieve precise lip synchronization, they involve substantial manual intervention and fail to synthesize full-head movements. In contrast, learning-based methods utilize neural networks to learn the mapping between audio and 3D facial motion in a data-driven manner, which have achieved remarkable progress in recent years.
However, all these methods generally rely on ground-truth-based errors for both training and evaluation, which may not align well with human perception.
Recently, \cite{chae2025perceptually} proposes to assess perceptual alignment between speech and lip movements from synchronization, readability, and expressiveness. However, they ignore other facial motions (\textit{e.g.}, upper-face and head movements) that also significantly influence perceptual quality. Moreover, their heuristic metrics developed without human annotations may face the same misalignment issues with human perception as other objective metrics \cite{sun2022deep,wu2022fast,wu2025fvq,wu2025hveval}.
In this paper, we comprehensively investigate the human perceptual alignment problem in audio-driven 3D facial animation for the first time, spanning human preference data collection, automatic perceptual modeling, and model optimization guided by perceptual quality.

\subsection{Learning from Human Feedback}
Generative models often fail to fully meet human intentions due to the limitations of objective or heuristic metrics in accurately reflecting human preferences. To address this issue, learning from human feedback \cite{christiano2017deep,ouyang2022training,xu2023imagereward} has gained increasing attention in recent years.
Such works typically learn reward models from human preference or rating annotations to approximate human perceptual judgments, and then incorporate these models into the training or evaluation process to improve the alignment of generated results with human preferences.
Specifically, reward models aim to automatically predict human-aligned perceptual scores from input signals, which has been widely studied in the image \cite{xu2023imagereward,gao2025multi} and video \cite{liu2025improving} domains, and recently extended to 3D content \cite{ye2024dreamreward,wang2025mvreward} and human motion \cite{voas2023best,wang2024aligning}. However, reward models for audio-driven 3D facial animation remain largely unexplored, and our FMReward is the first to fill this gap.
Early works \cite{ziegler2019fine,ouyang2022training} introduce the concept of learning from human feedback to language models and demonstrate its effectiveness in improving generation quality and alignment with user intentions.
More recently, this concept has been extended to text-to-image generation \cite{xu2023imagereward}, and subsequently expanded into text-to-3D generation \cite{ye2024dreamreward}, image-to-3D generation \cite{wang2025mvreward}, and text-to-motion generation \cite{wang2024aligning}. However, its application in audio-driven 3D facial animation remains unexplored.
To fill this gap, we collect the first human preference dataset for audio-driven 3D facial animation, build a reward model based on it, and employ it to improve the perceptual quality of the generated 3D facial motion.

\section{FMPair: Human Preference Dataset for Audio-driven 3D Facial Animation}

To alleviate the absence of human perceptual evaluation dataset for audio-driven 3D facial animation, we construct FMPair, the first pairwise preference dataset, consisting of 65,574 annotated 3D facial motion pairs from 8,834 distinct audio clips.

\begin{figure}[t]
\centering
\includegraphics[width=\linewidth]{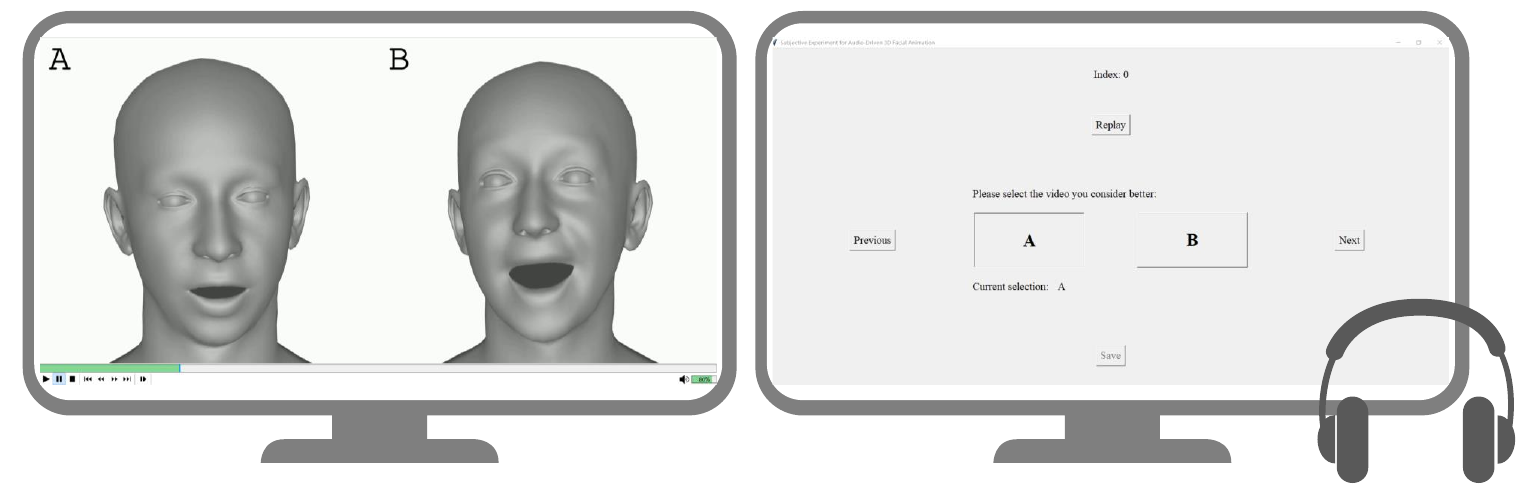}
\vspace{-2mm}
\caption{Illustration of the human preference annotation platform. Comparison videos are played in full screen and can be freely paused or sought.}
\vspace{-2mm}
\label{fig:gui}
\end{figure}

\subsection{Data Collection}
\label{sec: data_collect}
\noindent\textbf{Audio.}
The main principle of the audio collection process is to gather diverse audio clips that encompass human speeches in real-world scenarios.
To this end, we collect audio clips from two latest in-the-wild audio-driven 3D facial animation datasets. (1) TFHP \cite{sun2024diffposetalk}, which contains 26.5 hours of speech audios and corresponding 3D facial motion, derived from high-quality videos of 588 subjects in diverse real-world scenarios such as news, lectures, interviews, \textit{etc}. (2) MMHead \cite{wu2024mmhead}, which consists of 49 hours of audio and 3D facial motion, sourced from both YouTube videos captured in the wild and recordings collected in controlled laboratory environments, encompassing over 2,000 subjects.
We use the audio clips from the validation and test sets of these two datasets, and filter out clips shorter than 4 seconds.
As a result, we obtain a total of 8,834 audio clips, including 4,988 clips from TFHP \cite{sun2024diffposetalk} and 3,846 clips from MMHead \cite{wu2024mmhead}.

\noindent\textbf{3D Facial Motion.}
Based on the collected speech audio, we generate the corresponding 3D facial motions using four state-of-the-art audio-driven 3D facial animation methods, including DiffPoseTalk \cite{sun2024diffposetalk}, MM2Face \cite{wu2024mmhead}, UniTalker \cite{fan2024unitalker}, and ProbTalk3D \cite{wu2024probtalk3d}. The ground-truth facial motions from TFHP \cite{sun2024diffposetalk} and MMHead \cite{wu2024mmhead} are included as well.
To better reflect real-world scenarios, we require 3D facial motions that incorporate head poses. Given the lack of head poses in the motions generated by UniTalker \cite{fan2024unitalker} and ProbTalk3D \cite{wu2024probtalk3d}, we incorporate ground-truth head poses into the motions generated by these two methods.
To obtain diverse and abundant 3D facial motion sequences, we generate multiple motion sequences for each audio clip using non-deterministic methods, including DiffPoseTalk \cite{sun2024diffposetalk}, MM2Face \cite{wu2024mmhead}, and ProbTalk3D \cite{wu2024probtalk3d}.
To ensure temporal consistency, all motion sequences are standardized to 25 fps and aligned in duration with their corresponding audio. Specifically, for each audio clip, we compute a target frame length $N$ based on its duration and the fixed frame rate. Then, for each motion sequence associated with that audio clip, if its length exceeds $N$, it is truncated to the first $N$ frames; otherwise, it is padded to $N$ frames by repeating the last frame.
In addition, for MM2Face \cite{wu2024mmhead}, UniTalker \cite{fan2024unitalker}, and ProbTalk3D \cite{wu2024probtalk3d}, we employ their officially released pretrained models. For DiffPoseTalk \cite{sun2024diffposetalk}, we retrain a model without the style encoder while keeping all other settings unchanged.
In total, we obtain 229,684 distinct 3D facial motion sequences, which serve as the data pool for paired data construction.

\noindent\textbf{Pairwise Data.}
Based on the collected audio clips and corresponding 3D facial motion sequences, we construct pairwise data for human preference annotation.
Different from prior works \cite{xu2023imagereward,wang2024aligning} that include four samples per group for ranking, we present only two rendered videos in each group, as audio-driven 3D facial animations are relatively difficult to distinguish and fewer simultaneous comparisons help improve reliability.
We construct 3D facial motion pairs with and without head poses, motivated by two considerations: (1) most current audio-driven 3D facial animation methods do not generate head movements, such as UniTalker \cite{fan2024unitalker} and ProbTalk3D \cite{wu2024probtalk3d} we used; (2) excluding head movements can help annotators concentrate on facial dynamics.
Notably, for data without head poses, we add ground-truth head poses during the FMReward model training to avoid training bias.
Following the above principles, we ultimately constructed 65,574 3D facial motion pairs, including 36,105 pairs with head poses and 29,469 pairs without head poses. These pairs comprise 101,216 distinct motion sequences and 8,834 audio clips, with each pair sharing the same audio clip.

\begin{figure}[t]
\centering
\includegraphics[width=\linewidth]{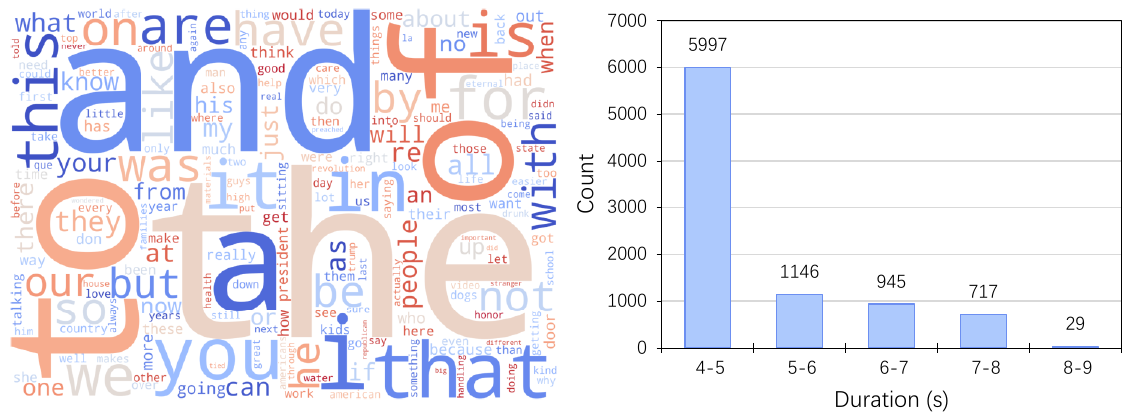}
\vspace{-2mm}
\caption{Overview of audio statistics. Left: word cloud of the audio transcripts. Right: bar chart of the audio length distribution.}
\vspace{-2mm}
\label{fig:audio}
\end{figure}

\subsection{Data Annotation}
\label{sec: data_anno}
\begin{figure*}
\centering
\includegraphics[width=\linewidth]{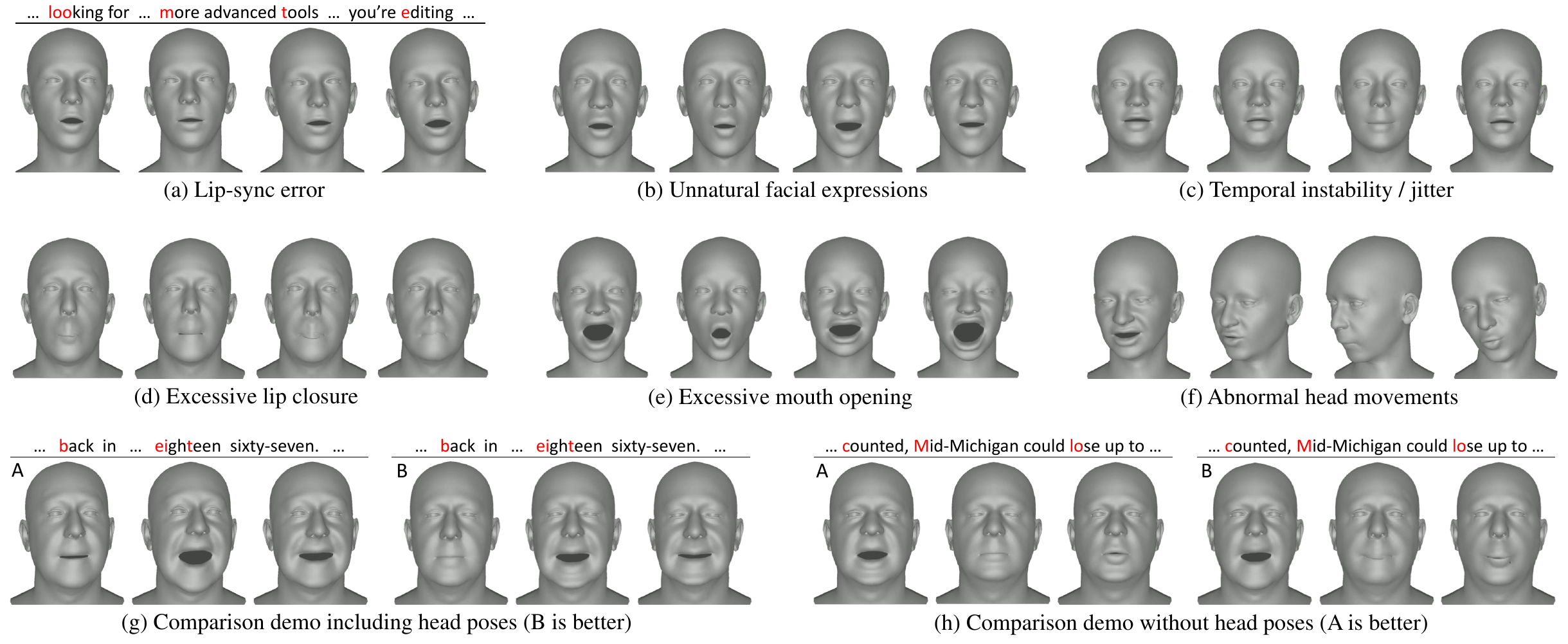}
\vspace{-3mm}
\caption{Representative examples of the FMPair dataset. (a)-(f) illustrate typical issues that may reduce the perceptual quality of 3D facial motion. (g) shows a pairwise comparison demo including head poses and its better choice, while (h) shows one without head poses.
}
\vspace{-2mm}
\label{fig:demo}
\end{figure*}

To collect human preference annotations for the pairwise data, we conduct subjective experiments following a carefully designed annotation pipeline.
For each data pair, annotators are required to select the 3D facial motion that they consider better according to the audio. If both motions are good, select the better one; if both are poor, select the comparatively better one.
Concretely, each annotator is first trained to become familiar with the rendered 3D facial motion and to identify potential factors that may lead to poor quality, as illustrated in Figure \ref{fig:demo}.
After the training session, a testing session with 10 data pairs is conducted to assess whether the annotators are sufficiently trained. Only those who pass the testing session are allowed to participate in the formal experiments.
In the formal experiments, annotators are asked to select their preferences using the platform shown in Figure \ref{fig:gui}. The comparison videos are played in full screen, and annotators are allowed to pause or replay them to make better judgments.
After all experiments, the final preference choices are obtained by majority voting across annotators.

\begin{figure}[t]
\centering
\includegraphics[width=0.85\linewidth]{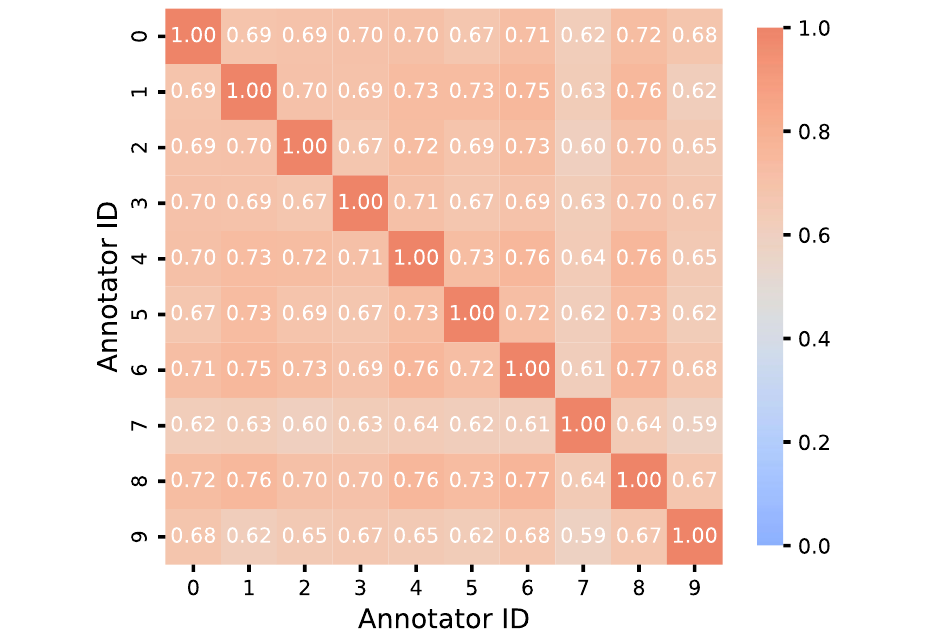}
\vspace{-2mm}
\caption{Heatmap of the pairwise agreement ratios among all annotators.}
\vspace{-2mm}
\label{fig:heatmap}
\end{figure}

\subsection{Data Analysis}
\label{sec: data_analysis}
Based on the aforementioned steps, we obtained a total of 65,574 annotated 3D facial motion pairs associated with 8,834 audio clips, which constitute our FMPair dataset.
In this section, we provide comprehensive analyses of the proposed FMPair dataset in terms of audio, 3D facial motion, and human preference annotations, respectively.

Firstly, Figure \ref{fig:audio} presents the word cloud of the speech text transcribed by Whisper \cite{radford2023robust} and the distribution of audio durations. It can be observed that the speech audio covers a broad vocabulary, especially high-frequency spoken words, and exhibits a rich variety of phonemes and diverse durations.
Secondly, Figure \ref{fig:demo} shows representative examples of 3D facial motion in the FMPair dataset, including typical issues and artifacts that may degrade the quality of 3D facial motion, as well as comparison cases with and without head poses. It is evident that common artifacts, such as unnatural facial expressions, temporal instability or jitter, excessive or abnormal mouth and head movements, cannot be properly identified or evaluated by existing ground-truth-based metrics, since audio to 3D facial motion mapping is inherently many-to-many while only a single ground truth is provided. This reveals the significance of human preferences in evaluating audio-driven 3D facial animation.
Finally, we calculate the pairwise agreement on 600 data pairs from 10 annotators and present the corresponding heatmap in Figure \ref{fig:heatmap}. It can be observed that annotators reach consensus on most data pairs, while some pairs are more dependent on individual human preferences. Therefore, the final choices are obtained by majority voting across annotators.

\section{FMReward: Automatic Perceptual Modeling to Capture Human Preferences}
\begin{figure*}[h]
\centering
\includegraphics[width=0.92\linewidth]{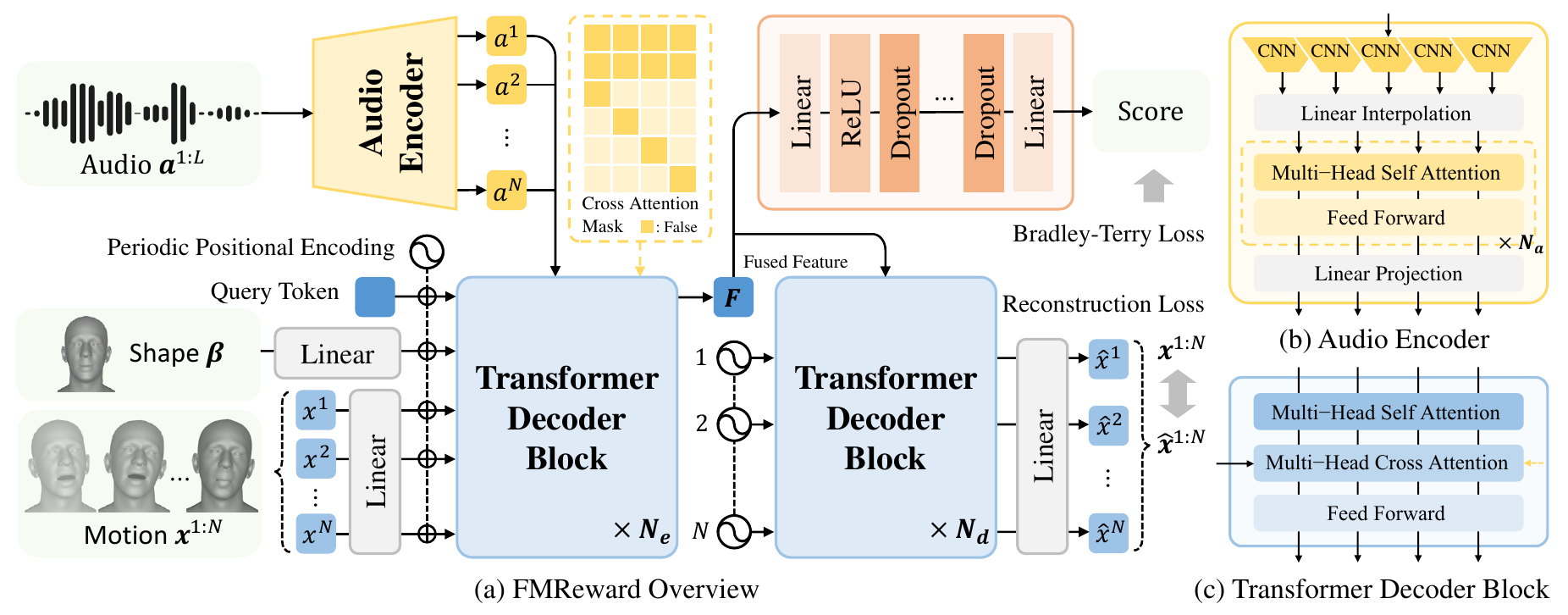}
\vspace{-2mm}
\caption{Overview of FMReward. Given 3D facial motion, FLAME shape parameters, and audio as inputs, motion and audio features are extracted by a transformer decoder and an audio encoder, fused via cross-attention, and passed through an MLP to predict a quality score. The training process takes a pair of samples as input, and supervises the fused features and predicted scores with the motion reconstruction loss and Bradley-Terry loss, respectively.}
\vspace{-2mm}
\label{fig:fmreward}
\end{figure*}

Considering the misalignment of existing metrics with human preferences and the labor-intensive nature of user studies, we propose FMReward, the first automatic perceptual model for audio-driven 3D facial animation, designed to predict quality scores that align with human perceptual judgments.
An overview of FMReward is illustrated in Figure \ref{fig:fmreward}.

\subsection{Architecture}
\label{sec：fmreward_arch}

\noindent\textbf{Problem Formulation.}
FMReward takes 3D facial motion $\boldsymbol{x}^{1:N}$, shape parameters $\boldsymbol{\beta}$, and audio $\boldsymbol{a}^{1:L}$ as inputs, and outputs a quality score $s$ that aligns with human preferences, which can be formulated as:
\begin{equation}
    s=\mathcal{R}(\boldsymbol{x}^{1:N},\boldsymbol{\beta},\boldsymbol{a}^{1:L}),
\label{eq:score}
\end{equation}
where $\mathcal{R}$ denotes the FMReward model. $N$ and $L$ are the unified sequence lengths of motion and audio after padding, where $L=f_a/f_v \times N$, with $f_a$ and $f_v$ denoting audio sampling rate and frame rate, respectively.
Specifically, the 3D facial motion is represented by 56-dimensional FLAME \cite{li2017learning} parameters, which can be formulated as $\boldsymbol{x}^{1:N}=\{ (\boldsymbol{\psi}^1, \boldsymbol{\theta}^1), ..., (\boldsymbol{\psi}^N, \boldsymbol{\theta}^N) \}$, where $\boldsymbol{\psi} \in \mathbb{R}^{50}$ and $\boldsymbol{\theta} \in \mathbb{R}^{6} $ denote FLAME expression and pose parameters, respectively. $\boldsymbol{\beta} \in \mathbb{R}^{100}$ corresponds to FLAME shape parameters, which characterize identity-related static facial geometry.

\noindent\textbf{Multimodal Feature Fusion.}
Audio and 3D facial motion feature extraction and fusion constitute the first step in predicting the quality score.
To achieve this, we employ a transformer decoder \cite{vaswani2017attention} architecture to fuse the two features through cross-attention.
In detail, the audio feature is extracted through an audio encoder $\mathcal{E}_a$ built upon the widely adopted speech model wav2vec 2.0 \cite{baevski2020wav2vec}, followed by a linear projection layer inspired by \cite{fan2022faceformer,wu2025singinghead}. The audio encoder is initialized using the pre-trained wav2vec 2.0 weights and fine-tuned during training.
The 3D facial motion feature is extracted by the transformer decoder $\mathcal{E}$, in which each motion frame and the shape parameters are first projected to the same dimension as the transformer latent space through two separate linear projection layers $\mathcal{P}_m$ and $\mathcal{P}_s$, and then encoded by self-attention layers.
Moreover, we include an additional learnable query token to extract the fused audio and 3D facial motion features. The query token is fed into the transformer decoder $\mathcal{E}$ together with the projected shape and motion tokens, and the corresponding position in the transformer output is taken as the fused feature $\boldsymbol{F}$:
\begin{equation}
    \boldsymbol{F} = \mathcal{E}(\text{PPE}(\mathcal{P}_m(\boldsymbol{x}^{1:N}), \mathcal{P}_s(\boldsymbol{\beta}), \mathcal{E}_a(\boldsymbol{a}^{1:L}))),
\end{equation}
where $\text{PPE}$ represents the periodic positional encoding \cite{fan2022faceformer} operation. $\boldsymbol{F} \in \mathbb{R}^{d}$, where $d$ is the transformer latent dimension. Notably, we adopt full self-attention without masking, allowing each token to attend to all others. As for cross-attention, we apply a hybrid attention mask \cite{sun2024diffposetalk,wu2024mmhead} as shown in Figure \ref{fig:fmreward}, in which the full-attention part allows the query token to aggregate information from all audio features, while the diagonal mask ensures temporal alignment between audio and motion and reduces interference from irrelevant frames.

\noindent\textbf{Motion Reconstruction.}
To enhance the representational capacity and robustness of the fused feature $\boldsymbol{F}$, we introduce a reconstruction branch to reconstruct the original 3D facial motion from the fused feature.
Specifically, we initialize $N$ temporal query tokens with zero vectors similar to \cite{petrovich2021action}, enrich them with periodic positional encoding \cite{fan2022faceformer}, and use them as inputs to a transformer decoder \cite{vaswani2017attention} for motion reconstruction. The fused feature that incorporates information from both audio and motion is fed into the transformer decoder through cross-attention without masking. Finally, the output features of the transformer are passed through a linear projection layer to obtain the reconstructed 3D facial motion $\boldsymbol{\hat{x}}^{1:N}$.

\noindent\textbf{Quality Regression.}
The quality regression module $\mathcal{Q}$ takes the fused feature $\boldsymbol{F}$ as input and outputs a scalar score $s$ that reflects perceptual quality, which can be formulated as $s=\mathcal{Q}(\boldsymbol{F})$.
Specifically, the module $\mathcal{Q}$ is a multilayer perceptron (MLP), which consists of four linear layers, each followed by ReLU activation and dropout, except for the final layer that directly outputs a scalar score.

\subsection{Training}
\label{sec：fmreward_train}
\noindent\textbf{Training Strategy.}
To train the FMReward model $\mathcal{R}$ using only pairwise comparison data without access to exact score values, we formulate the training objective as aligning the model’s pairwise judgments with human preferences, \textit{i.e.}, for any two samples, the one preferred by humans should also be assigned a higher score by the model.
To achieve this, we adopt the Bradley–Terry model \cite{bradley1952rank,wang2024aligning} following \cite{wang2024aligning}, which estimates the model parameters $\theta$ by maximizing the expected log-likelihood of observed human preferences:
\begin{equation}
\hat{\theta}=\arg\max_{\mathcal{\theta}} \mathbb{E}_{(i, j) \sim \mathcal{D}}  \left[ \log \sigma \big( s_i - s_j \big) \right],
\end{equation}
where $\mathcal{D}$ denotes the index set of the FMPair dataset, $i$ and $j$ correspond to two samples with humans preferring 3D facial motion $\boldsymbol{x}_i$ over $\boldsymbol{x}_j$, and $s_i$ and $s_j$ denote their quality scores assigned by the model $\mathcal{R}$. $\sigma$ represents the sigmoid function.

\noindent\textbf{Losses.}
The primary loss is the Bradley–Terry loss $\mathcal{L}_\text{BT}$ that forces the model judgments aligned with human preferences, which is formulated as:
\begin{equation}
    \mathcal{L}_\text{BT} = -\log \sigma(\mathcal{R}(\boldsymbol{x}_i,\boldsymbol{\beta},\boldsymbol{a})-\mathcal{R}(\boldsymbol{x}_j,\boldsymbol{\beta},\boldsymbol{a})),
\end{equation}
where $i$ corresponds to the better sample and $j$ to the worse one.
In addition, we add an auxiliary reconstruction loss $\mathcal{L}_\text{rc}$ to enhance the representational capacity and robustness of the fused feature $\boldsymbol{F}$, which can be written as:
\begin{equation}
    \mathcal{L}_\text{rc} = || \boldsymbol{\hat{x}}^{1:N} - \boldsymbol{x}^{1:N} ||_2^2,
\end{equation}
where $\boldsymbol{\hat{x}}^{1:N}$ and $\boldsymbol{x}^{1:N}$ represent the reconstructed and input 3D facial motion, respectively.
Notably, all loss terms are averaged across the batch.
The overall training loss is given by $\mathcal{L} = \mathcal{L}_\text{BT} + \lambda\mathcal{L}_\text{rc}$, with $\lambda$ denoting the trade-off coefficient between the two loss terms.

\begin{figure*}
\centering
\includegraphics[width=0.92\linewidth]{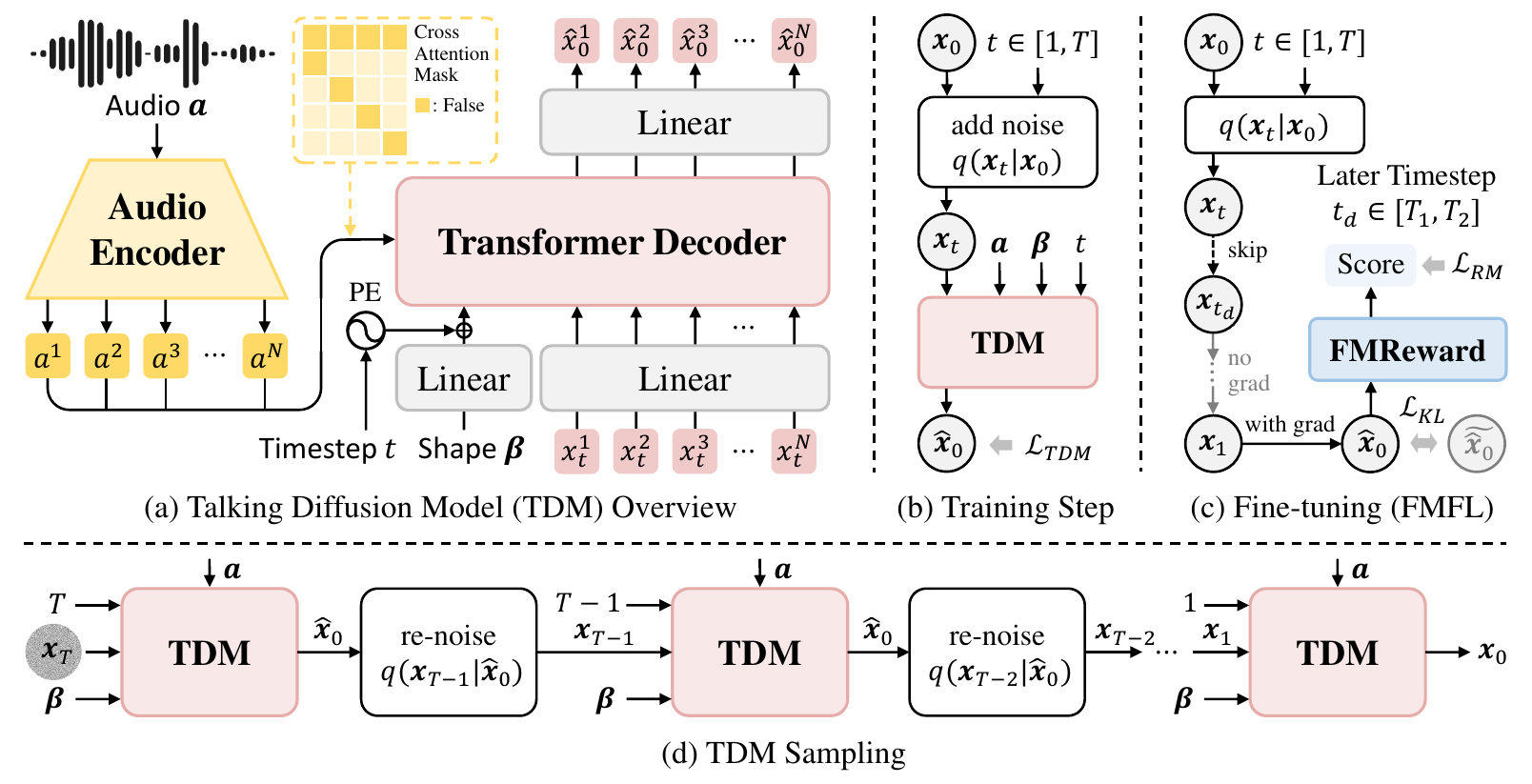}
\vspace{-2mm}
\caption{Overview of the diffusion-based framework for audio-driven 3D facial animation incorporating human preferences. (a) Denoising network of Talking Diffusion Model (TDM). Given 3D facial motion sequence $\boldsymbol{x}^{1:N}_t$ of length $N$ at denoising step $t$, FLAME shape parameters $\boldsymbol{\beta}$, timestep $t$, and audio $\boldsymbol{a}$, the transformer decoder predicts the clean motion $\boldsymbol{\hat{x}}^{1:N}_0$. (b) TDM training step. The ground-truth motion $\boldsymbol{x}_0$ is noised into $\boldsymbol{x}_t$ at a random step, which is then fed into the TDM for denoising and subsequently used to compute the training loss. (c) TDM fine-tuning step using FMFL. The noised motion $\boldsymbol{x}_t$ is first denoised for $t_d$ steps without gradients and then for one step with gradients, after which the result is fed into the reward model for supervision. (d) TDM sampling. After training and fine-tuning, TDM is able to generate 3D facial motion aligned with human preferences by iteratively denoising and re-noising from sampled random noise $\boldsymbol{x}_T$.
}
\vspace{-2mm}
\label{fig:fmfl}
\end{figure*}

\section{FMFL: Improving audio-driven 3D facial animation with FMReward}

Based on the FMReward model, it becomes feasible to improve the audio-driven 3D facial animation model for better alignment with human preferences.
To achieve this, we first construct a vanilla diffusion-based audio-driven 3D facial animation model (Section \ref{sec: fmfl_pre}), and then introduce FMFL, a direct fine-tuning algorithm that leverages the pretrained FMReward to optimize the diffusion model (Section \ref{sec: fmfl_fmfl}).

\subsection{Preliminary}
\label{sec: fmfl_pre}
We construct the Talking Diffusion Model (TDM), a well-performing vanilla diffusion-based audio-driven 3D facial animation model, based on DiffPoseTalk \cite{sun2024diffposetalk}, FaceDiffuser \cite{stan2023facediffuser}, and MDM \cite{tevet2022human}.
The TDM model is illustrated in Figure \ref{fig:fmfl}.
Given audio $\boldsymbol{a}^{1:L}$ and shape parameters $\boldsymbol{\beta}$, our TDM $\mathcal{T}$ can generate diverse 3D facial motion $\boldsymbol{x}^{1:N}$ represented by 56-dimensional FLAME \cite{li2017learning} parameters (\textit{i.e.}, 50 dimensions for expression and 6 dimensions for poses).

Specifically, TDM is built upon diffusion models \cite{sohl2015deep,ho2020denoising,song2020denoising}, which define a forward process that gradually perturbs data by adding Gaussian noise over a series of time steps, and a reverse process that learns to invert this corruption by gradually denoising from pure noise. In practice, the reverse process is approximated by a trainable denoising neural network. Once trained, the probabilistic model can generate target outputs by iteratively denoising Gaussian noise.
For TDM, the denoising network is implemented with a transformer decoder \cite{vaswani2017attention} architecture as shown in Figure \ref{fig:fmfl} (a). In detail, the denoising network takes noisy 3D facial motion sequence $\boldsymbol{x}^{1:N}_t$ at timestep $t$, FLAME shape parameters $\boldsymbol{\beta}$, timestep $t$, and audio $\boldsymbol{a}^{1:L}$ as inputs and outputs the clean motion $\boldsymbol{\hat{x}}^{1:N}_0$. The audio is encoded by an audio encoder implemented with HuBERT \cite{hsu2021hubert}, with an additional resampling layer \cite{sun2024diffposetalk} to ensure that the audio features $\boldsymbol{a}^{1:N}$ have the same sequence length as the motion sequence, which is then fed into the transformer decoder through cross-attention. The audio encoder is initialized using the pre-trained HuBERT \cite{hsu2021hubert} weights and fine-tuned during training. The 3D facial motion $\boldsymbol{x}^{1:N}_t$ and shape parameters $\boldsymbol{\beta}$ are projected to the transformer latent dimension via separate linear projection layers. The output motion $\boldsymbol{\hat{x}}^{1:N}_0$ is obtained by passing the transformer outputs, excluding the first token, through a linear projection layer.

To train TDM, we use the simple loss \cite{ramesh2022hierarchical,tevet2022human,sun2024diffposetalk} to supervise the predicted clean sample:
\begin{equation}
    \mathcal{L}_\text{simple} = || \boldsymbol{\hat{x}}^{1:N}_0 - \boldsymbol{x}^{1:N}_0 ||_2^2,
\end{equation}
where $\boldsymbol{\hat{x}}^{1:N}_0$ and $\boldsymbol{x}^{1:N}_0$ represent the predicted and ground-truth 3D facial motion, respectively. Auxiliary losses are further incorporated to supervise the predicted FLAME \cite{li2017learning} mesh vertices and head pose parameters, including MSE loss $\mathcal{L}_\text{MSE}$, velocity loss $\mathcal{L}_\text{vel}$, and smooth loss $\mathcal{L}_\text{smooth}$:
\begin{equation}
    \mathcal{L}_\text{MSE} = || \boldsymbol{\hat{X}}^{1:N} - \boldsymbol{X}^{1:N} ||_2^2,
\end{equation}
\begin{equation}
    \mathcal{L}_\text{vel} = ||(\boldsymbol{\hat{X}}^{2:T}-\boldsymbol{\hat{X}}^{1:T-1})- (\boldsymbol{X}^{2:T} - \boldsymbol{X}^{1:T-1})||_2^2,
\end{equation}
\begin{equation}
    \mathcal{L}_\text{smooth} = || \boldsymbol{\hat{X}}^{3:N} - 2\boldsymbol{\hat{X}}^{2:N-1} + \boldsymbol{\hat{X}}^{1:N-2} ||_2^2,
\end{equation}
where $\boldsymbol{\hat{X}}$ denotes the sequence of FLAME \cite{li2017learning} mesh vertices or pose parameters. Note that the FLAME mesh is obtained by forwarding the FLAME model with the head pose removed.
Based on the unified formulation, the overall auxiliary loss for vertices can be written as $\mathcal{L}_\text{vert}=\lambda_\text{m}\mathcal{L}_\text{MSE}+\lambda_\text{v}\mathcal{L}_\text{vel}+\lambda_\text{s}\mathcal{L}_\text{smooth}$, and the overall auxiliary loss for head pose, $\mathcal{L}_\text{pose}$, is defined in a similar way.
Finally, the overall training loss is defined as $\mathcal{L}_\text{TDM}=\mathcal{L}_\text{simple}+\mathcal{L}_\text{vert}+\mathcal{L}_\text{pose}$.

After training, we can generate diverse 3D facial motion sequences by sampling from Gaussian noise and iteratively denoising, as illustrated in Figure \ref{fig:fmfl} (d). Specifically, starting from a given diffusion step $T$ and randomly sampled noise $\boldsymbol{x}_T$, TDM gradually generates 3D facial motion by iteratively denoising and re-noising. At each timestep $t$, the denoising network predicts the clean sample $\boldsymbol{\hat{x}}_0$ from the noisy input $\boldsymbol{x}_t$, conditioned on the timestep $t$, audio $\boldsymbol{a}$, and shape parameters $\boldsymbol{\beta}$. The predicted clean sample $\boldsymbol{\hat{x}}_0$ is then re-noised to $\boldsymbol{x}_{t-1}$ for the subsequent step.

\subsection{FMFL}
\label{sec: fmfl_fmfl}
To improve the perceptual quality of 3D facial motion generated by TDM, we introduce the FMFL algorithm to fine-tune TDM with the guidance of pretrained perceptual model.

\begin{algorithm}[H]

\footnotesize

    \caption{\footnotesize FMFL: Fine-tuning TDM with FMReward}
    \begin{algorithmic}[1]
        \State \textbf{Dataset:} Audio-motion dataset $\mathcal{D} = \{ (\boldsymbol{a}_i, \boldsymbol{\beta}_i, \boldsymbol{m}_i)\}^n_{i=1}$
        \State \textbf{Input:} TDM model $\mathcal{T}$ with pretrained parameters $\theta_0$, TDM training loss function $\mathcal{L}_{\text{TDM}}$, reward model $\mathcal{R}$, reward-to-loss mapping function $\phi$, reward loss scale $\lambda_\text{r}$, and KL loss scale $\lambda_\text{k}.$
        \State \textbf{Initialization:} The number of diffusion step $T$, and the timestep range $[T_1, T_2]$ for fine-tuning.
        \For{ $(\boldsymbol{a}_i, \boldsymbol{\beta}_i, \boldsymbol{m}_i) \in \mathcal{D}$}
            \State $\theta_i \gets \theta_i$ \Comment{Update $\mathcal{T}_{\theta_i}$ using $\mathcal{L}_{\text{TDM}}$}
            \State $t \gets \text{rand}(1, T)$ \Comment{Pick a random timestep $t \in [1, T]$}
            \State $\boldsymbol{x}_t \sim q(\boldsymbol{x}_t \mid \boldsymbol{m}_i)$ \Comment{Sample $\boldsymbol{x}_t$}
            \State $t_d \gets \text{rand}(T_1, T_2)$ \Comment{Pick a random later timestep $t_d \in [T_1, T_2]$}
            \For{$j = t_d, ..., 2$}
                \State \textbf{no grad:} $\boldsymbol{x}_{j-1} \gets \mathcal{T}_{\theta_i}(\boldsymbol{x}_{j})$ 
            \EndFor
            \State \textbf{with grad:} ${\hat{\boldsymbol{x}}_0} \gets \mathcal{T}_{\theta_i}(\mathbf{x}_1)$
            \State $\mathcal{L}_{\text{RM}} \gets \phi(\mathcal{R}(\hat{\boldsymbol{x}}_0, \boldsymbol{\beta}_i, \boldsymbol{a}_i))$ \Comment{Reward loss}
            \State $\mathcal{L}_{\text{KL}} \gets \text{KL}(\widetilde{\hat{\boldsymbol{x}}_0}, \hat{\boldsymbol{x}}_0)$ \Comment{KL loss}
            \State $\theta_{i+1} \gets \theta_i$ \Comment{Update $\mathcal{T}_{\theta_i}$ with $\lambda_\text{r}\mathcal{L}_{\text{RM}} + \lambda_\text{k}\mathcal{L}_{\text{KL}}$}
            \State $\widetilde{\hat{\boldsymbol{x}}_0} \gets \hat{\boldsymbol{x}}_0$ \Comment{Update previous sample $\hat{\boldsymbol{x}}_0$}
        \EndFor
    \end{algorithmic}
    \label{table:refl}
\end{algorithm}

As illustrated in Algorithm 1, given an audio-driven 3D facial animation dataset $\mathcal{D}$ and pretrained reward model FMReward $\mathcal{R}$, we fine-tune TDM $\mathcal{T}_{\theta_0}$ trained on dataset $\mathcal{D}$ for better alignment with human preferences.
To ensure training stability and mitigate overfitting, we adopt a two-step update strategy at each fine-tuning iteration following \cite{xu2023imagereward,wang2024aligning}: (1) update TDM weights $\theta$ using its original training loss $\mathcal{L}_\text{TDM}$; (2) subsequently update TDM weights $\theta$ with the supervision of FMReward.
Specifically, the first step follows the TDM training procedure, where the ground-truth motion $\boldsymbol{x}_0$ is corrupted into $\boldsymbol{x}_t$ at a random timestep and then denoised by the denoising network to yield $\boldsymbol{\hat{x}}_0$ for loss computation, as illustrated in Figure \ref{fig:fmfl} (b).
In the second step, $\boldsymbol{x}_0$ is also randomly noised into $\boldsymbol{x}_t$, which is then denoised for $t_d$ steps without gradients and one step with gradients through DDIM sampling \cite{song2020denoising} to obtain $\boldsymbol{\hat{x}}_0$ for loss computation, as illustrated in Figure \ref{fig:fmfl} (c). In detail, $t_d \in [T_1,T_2]$ is a randomly selected later timestep. The predicted motion $\boldsymbol{\hat{x}}_0$, together with the input audio $\boldsymbol{a}$ and shape parameters $\boldsymbol{\beta}$, is fed into the reward model $\mathcal{R}$ to obtain a perceptual quality score, which is then used to compute the reward loss $\mathcal{L}_\text{RM}$:
\begin{equation}
    \mathcal{L}_\text{RM} = \phi(\mathcal{R}(\boldsymbol{\hat{x}}_0,\boldsymbol{\beta},\boldsymbol{a})),
\end{equation}
where $\phi(s)=\sigma(-s)$ is the reward-to-loss mapping function, and $\sigma$ represents the sigmoid function.
Moreover, considering that reward-based updates may shift the predicted motion distribution away from that learned during the original TDM training, we add a Kullback–Leibler (KL) divergence regularization $\mathcal{L}_\text{KL}$ to mitigate distributional drift between denoised samples across consecutive iterations during fine-tuning, thereby enhancing training stability. The KL loss is defined as:
\begin{equation}
\mathcal{L}_\text{KL}
= D_\text{KL}\!\left(
\mathcal{N}(\mu_{\text{p}}, \sigma_{\text{p}}^2)
\;\middle\|\;
\mathcal{N}(\mu_{\text{c}}, \sigma_{\text{c}}^2)
\right),
\end{equation}
\begin{equation}
D_{\text{KL}}
= 
\frac{1}{2}
\left[
\log\frac{\sigma_{\text{c}}^2}{\sigma_{\text{p}}^2}
+ \frac{\sigma_{\text{p}}^2 + (\mu_{\text{p}} - \mu_{\text{c}})^2}{\sigma_{\text{c}}^2}
- 1
\right],
\end{equation}
where $\mu_{\text{p}}$, $\sigma_{\text{p}}^2$ and $\mu_{\text{c}}$, $\sigma_{\text{c}}^2$ are means and variances of previous and current distributions derived from the predicted samples $\widetilde{\hat{\boldsymbol{x}}_0}$ and $\boldsymbol{\hat{x}}_0$ of the corresponding iterations, respectively.
The overall loss for the second step can be formulated as $\lambda_\text{r}\mathcal{L}_{\text{RM}} + \lambda_\text{k}\mathcal{L}_{\text{KL}}$, with $\lambda_\text{r}$ and $\lambda_\text{k}$ denoting the weighting coefficients.

\section{Experiments}

\begin{table*}[h]
\centering
\renewcommand\arraystretch{1}
\caption{Quantitative comparisons of 3D facial motion evaluation methods in terms of alignment with human preferences on the proposed FMPair dataset and its TFHP and MMHead subsets. \raisebox{0.2ex}{\scalebox{0.8}{$\diamondsuit$}}, \raisebox{0.2ex}{\scalebox{0.8}{$\clubsuit$}}, \raisebox{0.2ex}{\scalebox{0.8}{$\heartsuit$}}, and \raisebox{0.2ex}{\scalebox{0.8}{$\spadesuit$}} denote common audio-driven 3D facial animation metrics, audio-video alignment methods, audio-visual quality assessment methods, and video quality assessment methods, respectively. The abbreviations Acc., BS, CE, and $\rho$ represent Accuracy, Brier Score, Cross-Entropy, and SRCC, respectively. The best and runner-up performances are bold and underlined, respectively.}

\resizebox{\linewidth}{!}{
\begin{tabular}{lcccccccccccc}
\toprule
Dataset  & \multicolumn{4}{c}{TFHP Subset}  & \multicolumn{4}{c}{MMHead Subset}  & \multicolumn{4}{c}{FMPair} \\ \cmidrule(lr){2-5} \cmidrule(lr){6-9} \cmidrule(lr){10-13} 

Method / Metric  & Acc.\,$\uparrow$  & BS\,$\downarrow$  & CE\,$\downarrow$  & $\rho$\,$\uparrow$  & Acc.\,$\uparrow$  & BS\,$\downarrow$  & CE\,$\downarrow$  & $\rho$\,$\uparrow$  & Acc.\,$\uparrow$  & BS\,$\downarrow$  & CE\,$\downarrow$  & $\rho$\,$\uparrow$ \\
\midrule

\raisebox{0.2ex}{\scalebox{0.8}{$\diamondsuit$}} LVE	&	19.72\%		&	0.2516 	&	0.6964 	&	-0.70 	&	82.68\%	&	0.2490 	&	0.6912 	&	\underline{0.70} 	&	35.68\%	&	0.2510 	&	0.6951 	&	-0.10 	\\
\raisebox{0.2ex}{\scalebox{0.8}{$\diamondsuit$}} FVE	&	9.09\%		&	0.2504 	&	0.6940 	&	-0.70 	&	79.19\%	&	0.2498 	&	0.6928 	&	0.60 	&	26.87\%	&	0.2503 	&	0.6937 	&	-0.30 	\\
\raisebox{0.2ex}{\scalebox{0.8}{$\diamondsuit$}} FDD	&	20.87\%		&	0.2500 	&	0.6932 	&	-0.90 	&	51.29\%	&	0.2500 	&	0.6931 	&	0.10 	&	28.59\%	&	0.2500 	&	0.6931 	&	-0.50 	\\
\raisebox{0.2ex}{\scalebox{0.8}{$\diamondsuit$}} PE	&	5.01\%		&	0.9143 	&	7.2396 	&	-1.00 	&	71.05\%	&	0.1906 	&	0.5648 	&	0.20 	&	21.75\%	&	0.7308 	&	5.5469 	&	-0.90 	\\
\raisebox{0.2ex}{\scalebox{0.8}{$\diamondsuit$}} BA	&	23.61\%		&	0.2535 	&	0.7003 	&	-0.50 	&	32.62\%	&	0.2516 	&	0.6965 	&	-0.30 	&	25.89\%	&	0.2531 	&	0.6993 	&	-0.60 	\\
\hdashline

\raisebox{0.2ex}{\scalebox{0.8}{$\clubsuit$}} SyncNet-C \cite{chung2016out}	&	\underline{90.31\%}		&	\underline{0.0679} 	&	\underline{0.2320} 	&	\underline{0.80} 	&	\underline{86.35\%}	&	\underline{0.0897} 	&	\underline{0.2934} 	&	\textbf{0.90} 	&	\underline{89.30\%}	&	\underline{0.0734} 	&	\underline{0.2476} 	&	\textbf{1.00} 	\\
\raisebox{0.2ex}{\scalebox{0.8}{$\clubsuit$}} SyncNet-D \cite{chung2016out}	&	9.95\%		&	0.6982 	&	2.3501 	&	-0.90 	&	15.72\%	&	0.6311 	&	2.0309 	&	-0.90 	&	11.41\%	&	0.6812 	&	2.2692 	&	-1.00 	\\
\raisebox{0.2ex}{\scalebox{0.8}{$\clubsuit$}} VAST \cite{chen2023vast}	&	67.01\%		&	0.2499 	&	0.6930 	&	-0.10 	&	62.87\%	&	0.2499 	&	0.6930 	&	-0.10 	&	65.96\%	&	0.2499 	&	0.6930 	&	0.50 	\\
\raisebox{0.2ex}{\scalebox{0.8}{$\clubsuit$}} ImageBind \cite{girdhar2023imagebind}	&	53.68\%		&	0.2476 	&	0.6883 	&	0.70 	&	51.68\%	&	0.2494 	&	0.6920 	&	0.10 	&	53.18\%	&	0.2480 	&	0.6892 	&	0.30 	\\
\hdashline

\raisebox{0.2ex}{\scalebox{0.8}{$\heartsuit$}} ANNAVQA \cite{cao2023attention}	&	47.94\%		&	0.2502 	&	0.6936 	&	0.20 	&	43.48\%	&	0.2506 	&	0.6943 	&	0.20 	&	46.81\%	&	0.2503 	&	0.6937 	&	-0.40 	\\
\raisebox{0.2ex}{\scalebox{0.8}{$\heartsuit$}} GeneralAVQA \cite{cao2023subjective}	&	62.58\%		&	0.2490 	&	0.6912 	&	\underline{0.80} 	&	66.99\%	&	0.2490 	&	0.6912 	&	0.60 	&	63.70\%	&	0.2490 	&	0.6912 	&	\underline{0.60} 	\\
\hdashline

\raisebox{0.2ex}{\scalebox{0.8}{$\spadesuit$}} SimpleVQA \cite{sun2022deep}	&	54.00\%		&	0.2482 	&	0.6896 	&	0.30 	&	53.79\%	&	0.2483 	&	0.6898 	&	0.00 	&	53.95\%	&	0.2482 	&	0.6896 	&	-0.20 	\\
\raisebox{0.2ex}{\scalebox{0.8}{$\spadesuit$}} FastVQA \cite{wu2022fast}	&	59.49\%		&	0.2444 	&	0.6819 	&	\underline{0.80} 	&	62.30\%	&	0.2420 	&	0.6772 	&	\underline{0.70} 	&	60.20\%	&	0.2438 	&	0.6807 	&	0.20 	\\
\raisebox{0.2ex}{\scalebox{0.8}{$\spadesuit$}} Dover \cite{wu2023exploring}	&	42.02\%		&	0.4985 	&	2.6240 	&	0.30 	&	50.36\%	&	0.4119 	&	1.9090 	&	0.20 	&	44.13\%	&	0.4765 	&	2.4427 	&	-0.10 	\\
\hdashline

\rowcolor[gray]{.92}
\textbf{FMReward (Ours)}	&	\textbf{95.36\%}		&	\textbf{0.0299} 	&	\textbf{0.0945} 	&	\textbf{0.90} 	&	\textbf{90.11\%}	&	\textbf{0.0616} 	&	\textbf{0.1864} 	&	\textbf{0.90} 	&	\textbf{94.03\%}	&	\textbf{0.0379} 	&	\textbf{0.1178} 	&	\textbf{1.00} 	\\

\bottomrule
\end{tabular}
}
\label{tab:fmreward_compare}
\end{table*}

\begin{figure*}[h]
\centering
\includegraphics[width=\linewidth]{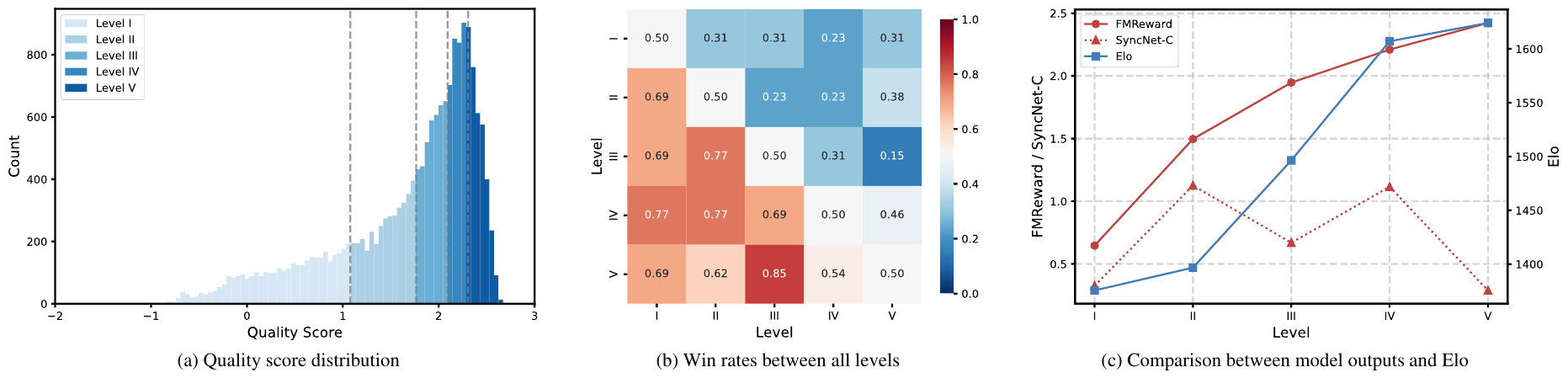}
\caption{Experiments on in-the-wild data. (a) Histogram and quality level division of quality scores predicted by FMReward. (b) Heatmap of win rates between all quality levels based on the user study. (c) Comparison of average quality scores across all levels among our FMReward, the runner-up method SyncNet-C, and the Elo ratings estimated from pairwise user preferences.}
\label{fig:exp_rm}
\end{figure*}

\subsection{Implementation details}
\noindent\textbf{FMReward.}
We train and evaluate the proposed FMReward model on the pairwise comparison dataset FMPair. The FMPair dataset is randomly divided into training and test sets with a ratio of 80\% : 20\%, resulting in 52,459 pairs for training and 13,115 pairs for testing.
The two transformer decoders are implemented as 4-layer transformer decoders (\textit{i.e.}, $N_e=N_d=4$), with a latent dimension of 256, 4 attention heads, a feed-forward size of 1024, GELU activation, and a dropout rate of 0.1.
The hidden dimensions of the quality regression MLP are 256, 128, 64, 32, 1.
The weighting coefficient $\lambda$ for the training loss is set to 0.1.
We train the model using the AdamW optimizer with learning rate $\eta=1 \times 10^{-4}$ and $\beta=(0.9, 0.999)$.
The model is trained for 10 epochs with a batch size of 48, which takes about 1.5 hours on 4 NVIDIA 4090 GPUs.

\noindent\textbf{TDM.}
We train the TDM model on the training set of the TFHP dataset \cite{sun2024diffposetalk}, which contains about 20 hours of speech audio and corresponding 3D facial motion sequences.
Following DiffPoseTalk \cite{sun2024diffposetalk}, all sequences are segmented into clips of 100 frames (4s), and two consecutive clips are used during training to enable seamless transitions when generating long sequences.
The diffusion step $T$ is set to 500.
The weighting coefficients are set to $\lambda_\text{m}=2 \times 10^6$, $\lambda_\text{v}=1 \times 10^7$, $\lambda_\text{s}=1 \times 10^5$ for $\mathcal{L}_\text{vert}$, and $\lambda_\text{m}=0.05$, $\lambda_\text{v}=5$, $\lambda_\text{s}=0.5$ for $\mathcal{L}_\text{pose}$.
We train the model using the Adam optimizer with $\beta=(0.9, 0.999)$. The learning rate is linearly warmed up to $\eta=1 \times 10^{-4}$ within the first 50k iterations and then maintained thereafter.
The model is trained for 110k iterations with a batch size of 16, which takes about 10 hours on 1 NVIDIA 4090 GPU.

\noindent\textbf{FMFL.}
We introduce the FMFL algorithm to fine-tune the pretrained TDM model on the TFHP \cite{sun2024diffposetalk} training set, leveraging the pretrained FMReward model to achieve better alignment with human preferences.
The later timestep range $[T_1,T_2]$ is set to $[350,450]$. The weighting coefficients are set to $\lambda_\text{r}=10$ and $\lambda_\text{k}=1$.
We fine-tune the model using the Adam optimizer with learning rate $\eta=1 \times 10^{-6}$ and $\beta=(0.9, 0.999)$. The model is fine-tuned for 240 iterations with a batch size of 24, which takes about 3.5 hours on 1 NVIDIA 4090 GPU.

\subsection{Evaluation of FMReward}
\noindent\textbf{Evaluation Metrics.}
We evaluate FMReward’s ability to predict quality scores aligned with human preferences from both pairwise comparison and absolute score prediction perspectives.
For the pairwise comparison evaluation, we convert predicted scores into pairwise comparisons and compare them with ground-truth human preference pairs using Accuracy, Brier Score, and Cross-Entropy. Accuracy provides the most direct measure of pairwise comparison correctness, which is defined as the proportion of correctly predicted pairs among all annotated pairs. Brier Score and Cross-Entropy further evaluate the quality of probabilistic predictions. Concretely, the Brier Score measures the mean squared difference between the predicted probabilities $p_i$ and the true labels $y_i$, which can be formulated as $\frac{1}{N}\sum_{i=1}^{N}(p_i - y_i)^2$. The Cross-Entropy quantifies the divergence between the predicted probabilities $p_i$ and the true labels $y_i$, imposing stronger penalties on overconfident incorrect predictions, and is formulated as $-\frac{1}{N}\sum_{i=1}^{N}\big[y_i\log(p_i) + (1 - y_i)\log(1 - p_i)\big]$. The predicted probabilities $p_i$ are obtained by performing the softmax function on the predicted quality score pairs. Together, these metrics ensure that the model not only makes correct pairwise decisions but also produces well-calibrated and reliable confidence estimates.
For the score prediction evaluation, we evaluate whether the predicted quality scores can correctly rank 3D facial animation methods according to human perceptual judgments, using Spearman’s Rank Correlation Coefficient (SRCC) as the evaluation metric. Specifically, SRCC is computed at the method level using the average predicted scores and the corresponding human preference scores for each method, where the latter are estimated from ground-truth pairwise comparison data using the Bradley–Terry model \cite{bradley1952rank}.

\begin{table}[t]
\centering
\renewcommand\arraystretch{1}
\caption{Ablation study of FMReward on the FMPair dataset. The abbreviations Acc., BS, CE, and $\rho$ represent Accuracy, Brier Score, Cross-Entropy, and SRCC, respectively.}
\resizebox{0.88\linewidth}{!}{
\begin{tabular}{lcccc}
\toprule

Method / Metric  & Acc.\,$\uparrow$  & BS\,$\downarrow$  & CE\,$\downarrow$  & $\rho$\,$\uparrow$ \\ 
\midrule

w/o $\mathcal{L}_\text{rc}$	&	93.76\%	&	0.0382 	&	0.1182 	&	0.90 	\\
w/o $\mathcal{L}_\text{BT}$	&	37.83\%	&	0.2512 	&	0.6955 	&	-0.40 	\\
w/o Shape Input	&	93.95\%	&	0.0381 	&	0.1179 	&	1.00 	\\
w/o Audio Input	&	93.79\%	&	0.0385 	&	0.1195 	&	1.00 	\\
FMReward-Small	&	93.85\%	&	0.0382 	&	0.1192 	&	1.00 	\\
FMReward-Large	&	93.60\%	&	0.0428 	&	0.1311 	&	1.00 	\\
\rowcolor[gray]{.92}
\textbf{FMReward (Ours)}	&	\textbf{94.03\%}	&	\textbf{0.0379} 	&	\textbf{0.1178} 	&	\textbf{1.00} 	\\

\bottomrule
\end{tabular}
}
\vspace{-3mm}
\label{tab:fmreward_ablation}
\end{table}

\noindent\textbf{Quantitative Comparison.}
We compare our model with commonly used audio-driven 3D facial animation metrics \cite{richard2021meshtalk,wu2025singinghead,xing2023codetalker,siyao2022bailando} and state-of-the-art audio-video alignment methods \cite{chung2016out,chen2023vast,girdhar2023imagebind}, audio-visual quality assessment methods \cite{cao2023attention,cao2023subjective}, and video quality assessment methods \cite{sun2022deep,wu2022fast,wu2023exploring}.
Specifically, commonly used audio-driven 3D facial animation metrics include Lip Vertex Error (LVE), Face Vertex Error (FVE), Upper-face Dynamics Deviation (FDD), Parameter Error (PE), and Beat Align Score (BA). LVE and FVE measure vertex-level reconstruction errors with respect to ground-truth facial motions, FDD evaluates the temporal dynamics of upper-face movements, PE assesses the accuracy of generated FLAME motion parameters, and BA quantifies the alignment between generated head motion and audio beats. In particular, PE is defined as the L2 distance between generated and ground-truth FLAME parameter vectors, while all other metrics follow their original definitions.
As shown in Table \ref{tab:fmreward_compare}, current 3D facial animation metrics fail to produce scores that align well with human preferences, which underscores the importance of developing perceptually aligned evaluation models.
Meanwhile, existing audio-visual \cite{cao2023attention,cao2023subjective
} and video \cite{sun2022deep,wu2022fast,wu2023exploring} quality assessment models perform poorly on 3D facial motion, largely due to their lack of direct access to 3D motion data and inadequate modeling of audio–motion synchronization.
General-purpose audio-visual alignment methods \cite{chen2023vast,girdhar2023imagebind} exhibit similar limitations and thus also perform poorly.
The confidence score of SyncNet \cite{chung2016out} achieves a relatively better alignment with human preferences, benefiting from its focus on human faces and explicit modeling of lip synchronization.  
In comparison, the proposed FMReward method achieves the best performance across all evaluation metrics, demonstrating its ability to automatically evaluate audio-driven 3D facial animation in a way that closely reflects human perceptual judgments.

\noindent\textbf{User Study.}
We further conduct a user study to evaluate the human perceptual-aligned quality score prediction capability of FMReward.
To evaluate the generalization ability of FMReward on in-the-wild data beyond the proposed FMPair dataset, we utilize the training set of the TFHP \cite{sun2024diffposetalk} dataset, which contains identities that do not appear in the FMPair dataset.
We first obtain a quality score for each motion using FMReward and visualize the score distribution in Figure \ref{fig:exp_rm} (a).
Then, we divide the data into five quality levels according to the quality scores, ensuring that each level contains an equal number of samples.
In the user study, participants are asked to compare randomly sampled 3D facial motion sequences with audio from each quality level, and the final preference for each pair is determined by majority voting. The user study involves eleven participants. The win rates between all quality levels are shown in Figure \ref{fig:exp_rm} (b), which indicate how often motions from one level are preferred over another. Motions from higher quality levels receive clearly more preferences than those from lower levels, demonstrating good alignment between our model’s predicted scores and human perceptual judgments.
Moreover, to enable direct comparison between the predicted scores of different metrics and human judgments, we estimate Elo ratings \cite{elo1978rating} for each quality level from pairwise comparison data obtained in the user study, and compare them with the scores obtained by our FMReward model and the runner-up model SyncNet-C \cite{chung2016out}.
Specifically, we treat each quality level as a ``player” and each human preference comparison as a match outcome. The Elo ratings are initialized equally and iteratively updated based on pairwise comparison results, where the preferred sample is treated as the winner. To improve stability, we randomly shuffle the pairwise comparison results multiple times and average the resulting ratings.
As shown in Figure \ref{fig:exp_rm} (c), the trend of our model's predictions matches the Elo ratings well, whereas SyncNet-C does not, further demonstrating the superiority of our model.

\noindent\textbf{Ablation Study.}
To validate the effectiveness of each component in our FMReward method, we conduct comprehensive ablation studies, as shown in Table \ref{tab:fmreward_ablation}.
First, we conduct ablation studies on the loss functions to verify the effectiveness of both the Bradley–Terry loss $\mathcal{L}_\text{BT}$ and the reconstruction loss $\mathcal{L}_\text{rc}$. In particular, $\mathcal{L}_\text{BT}$ plays a crucial role during training, as it forces the predicted scores to yield the same pairwise comparisons as human preferences.
Second, we demonstrate the effectiveness of providing additional information (\textit{i.e.}, audio and FLAME shape parameters) beyond 3D facial motion, as shown in the third and fourth rows of Table \ref{tab:fmreward_ablation}.
Finally, we investigate the impact of model size by defining three architectural variants of FMReward with different numbers of transformer parameters. Specifically, FMReward-Small is configured with a latent hidden dimension of 128, a feed-forward network hidden size of 512, 2 attention heads, and 2 transformer decoder layers. FMReward-Medium (Ours) and FMReward-Large increase these settings to 256, 1024, 4, 4, and 512, 2048, 8, 8, respectively. The performance differences among these models are moderate, with the medium-sized model achieving the best results.

\begin{table}
\centering
\renewcommand\arraystretch{1}
\caption{Comparison of 3D facial animation metrics at different fine-tuning iterations. The units of LVE and FDD are $mm$ and $\times 10^{-2} mm$, respectively. The best and runner-up performances are bold and underlined, respectively.}
\resizebox{\linewidth}{!}{
\begin{tabular}{cccccc}
\toprule

Iteration  & LVE.\,$\downarrow$  & FDD\,$\downarrow$  & Diversity\,$\rightarrow$  & BA\,$\uparrow$  & FMReward\,$\uparrow$ \\ 
\midrule

0	&	9.7750 	&	\textbf{0.5352} 	&	26.088 	&	0.2275 	&	1.0453 	\\
20	&	10.1472 	&	0.5801 	&	27.680 	&	0.2299 	&	1.4765 	\\
40	&	10.1349 	&	0.5711 	&	27.696 	&	0.2291 	&	2.1445 	\\
60	&	10.0560 	&	0.5560 	&	27.146 	&	0.2293 	&	2.3061 	\\
80	&	9.9478 	&	\underline{0.5484} 	&	26.522 	&	0.2310 	&	2.3451 	\\
100	&	9.8885 	&	0.5501 	&	26.205 	&	0.2308 	&	2.3596 	\\
120	&	9.8504 	&	0.5536 	&	26.035 	&	0.2301 	&	2.3724 	\\
140	&	9.8111 	&	0.5552 	&	25.807 	&	0.2323 	&	2.3851 	\\
160	&	9.7882 	&	0.5580 	&	25.672 	&	0.2316 	&	2.3946 	\\
180	&	9.7487 	&	0.5613 	&	25.560 	&	0.2308 	&	2.4037 	\\
200	&	9.7366 	&	0.5625 	&	25.539 	&	0.2308 	&	2.4066 	\\
220	&	9.7412 	&	0.5618 	&	25.580 	&	\textbf{0.2337} 	&	2.4045 	\\
\rowcolor[gray]{.92}
240	&	\underline{9.7264} 	&	0.5655 	&	25.412 	&	0.2322 	&	\textbf{2.4115} 	\\
260	&	9.7293 	&	0.5685 	&	25.396 	&	0.2308 	&	\underline{2.4112} 	\\
280	&	9.7383 	&	0.5724 	&	25.432 	&	0.2313 	&	2.4085 	\\
300	&	\textbf{9.7239} 	&	0.5734 	&	25.303 	&	\underline{0.2325} 	&	2.4106 	\\

\bottomrule
\end{tabular}
}
\label{tab:fmfl_iter}
\end{table}

\subsection{Evaluation of FMFL}

\begin{figure*}[h]
\centering
\includegraphics[width=0.92\linewidth]{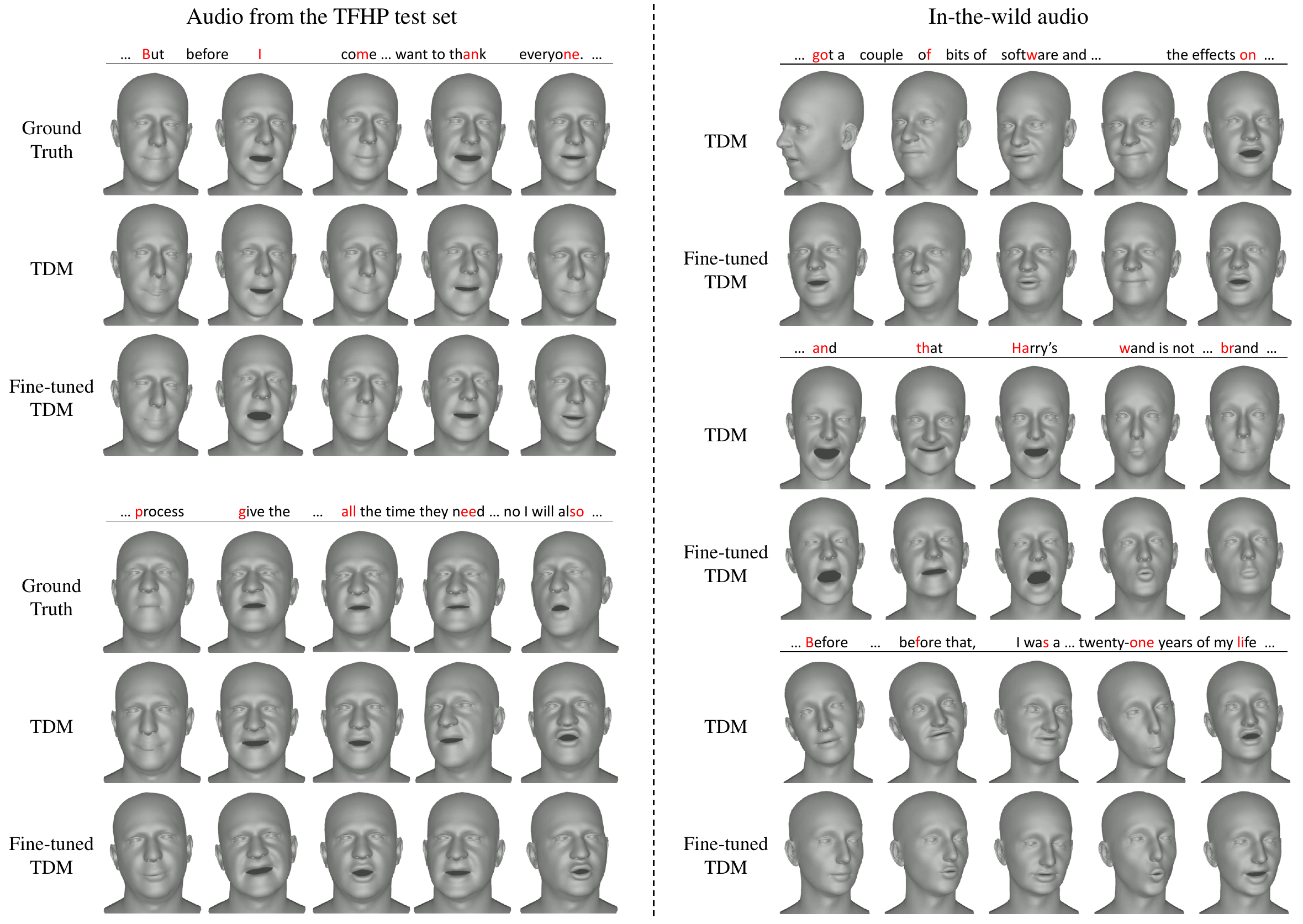}
\caption{Qualitative comparison before and after fine-tuning with FMFL. The left side shows 3D facial animation results driven by audio from the TFHP \cite{sun2024diffposetalk} test set, while the right side shows results driven by in-the-wild audio.
}
\vspace{-1mm}
\label{fig:fmfl_vis}
\end{figure*}

\noindent\textbf{Evaluation Metrics.}
We evaluate the effectiveness of the FMFL algorithm by assessing the performance of audio-driven 3D facial animation after fine-tuning with FMFL from four perspectives.
(1) Facial accuracy.
We employ two commonly used metrics, \textit{i.e.}, Lip Vertex Error (LVE) \cite{richard2021meshtalk} and Upper-face Dynamics Deviation (FDD) \cite{xing2023codetalker}, to evaluate the accuracy of generated facial movements against the ground truth while keeping the head fixed. LVE measures the lip synchronization by calculating the average value of the maximal $L_2$ error of all lip vertices in each frame. FDD focuses on upper face dynamics and is defined as the difference in standard deviation of upper-face vertices over time between generated and ground-truth sequences.
(2) Head motion.
We measure the alignment between the generated head motion and the input audio by applying the Beat Align Score (BA) \cite{siyao2022bailando} to the generated 3-dimensional head pose parameters.
(3) Diversity.
We evaluate the generation diversity by computing the average pairwise $L_2$ distance among generated samples, which can be formulated as $\frac{1}{n(n-1)} {\textstyle \sum_{i=1}^{n}} {\textstyle \sum_{j\ne i}^{n}} ||\boldsymbol{x}_i-\boldsymbol{x}_j||_2^2$, where $\boldsymbol{x}_i$ and $\boldsymbol{x}_j$ are generated 56-dimensional FLAME \cite{li2017learning} parameters.
(4) Human perception.
To quantify the perceptual quality of the generated 3D facial motion, we employ our pretrained FMReward to obtain a human preferences-aligned quality score as an evaluation metric (see Equation \ref{eq:score}).

\begin{table}[t]
\centering
\renewcommand\arraystretch{1}
\caption{Quantitative comparisons of audio-driven 3D facial animation methods on the TFHP \cite{sun2024diffposetalk} test set. ``Ours" refers to the TDM model fine-tuned with FMFL. The units of LVE and FDD are $mm$ and $\times 10^{-2} mm$, respectively. The best and runner-up performances are bold and underlined, respectively.}
\resizebox{\linewidth}{!}{
\begin{tabular}{lccccc}
\toprule

Method  & LVE.\,$\downarrow$  & FDD\,$\downarrow$  & Diversity\,$\rightarrow$  & BA\,$\uparrow$  & FMReward\,$\uparrow$ \\ 
\midrule

FaceFormer \cite{fan2022faceformer}	&	\underline{9.8246} 	&	0.8592 	&	0 	&	\textbf{0.2452} 	&	2.0182 	\\
CodeTalker \cite{xing2023codetalker}	&	13.413 	&	2.1878 	&	0 	&	0.2155 	&	1.9630 	\\
FaceDiffuser \cite{stan2023facediffuser}	&	17.563 	&	1.0429 	&	32.465 	&	0.2307 	&	0.9166 	\\
DiffPoseTalk \cite{sun2024diffposetalk}	&	9.9550 	&	\underline{0.6581} 	&	32.507 	&	0.2217 	&	\underline{2.1159} 	\\
\rowcolor[gray]{.92}
\textbf{Ours}  &	\textbf{9.7264} 	&	\textbf{0.5655} 	&	25.412 	&	\underline{0.2322} 	&	\textbf{2.4115} 	\\


\bottomrule
\end{tabular}
}
\label{tab:fmfl_baseline}
\end{table}

\begin{figure}
\centering
\includegraphics[width=0.9\linewidth]{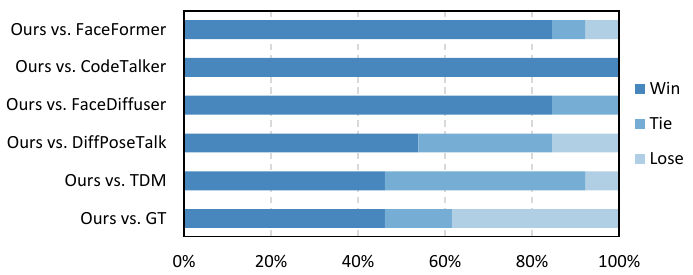}
\vspace{-1mm}
\caption{User study for audio-driven 3D facial animation. We show the win rates of our model (TDM fine-tuned with FMFL) compared to other models.}
\vspace{-3mm}
\label{fig:fmfl_user_study}
\end{figure}

\noindent\textbf{Quantitative Comparison.}
We first report the evolution of all evaluation metrics throughout the FMFL-based fine-tuning process, as shown in Table \ref{tab:fmfl_iter}.
It can be observed that the human perceptual score produced by FMReward steadily increases during fine-tuning, demonstrating the effectiveness of the FMFL algorithm. Meanwhile, BA and LVE either increase directly or exhibit a slight initial decrease followed by a clear upward trend, indicating that the fine-tuning process does not conflict with other metrics but instead jointly enhances most of them. Moreover, only a few hundred fine-tuning iterations are needed to improve the perceptual quality of the generated 3D facial motion, whereas the training process requires over 100k iterations.
We select the model that achieves the best FMReward score, namely the one fine-tuned for 240 iterations, as our final model, which is then used for comparison against other audio-driven 3D facial animation methods.
As shown in Table \ref{tab:fmfl_baseline}, our model achieves promising performance across all evaluation metrics, particularly showing higher perceptual quality and more accurate lip and upper-face movements. Note that mesh-based baseline methods \cite{fan2022faceformer,xing2023codetalker} are modified to use FLAME \cite{li2017learning} parameters for a fair comparison.

\noindent\textbf{Qualitative Comparison.}
We also provide a qualitative comparison before and after fine-tuning TDM with the FMFL algorithm. Both test-set audio and in-the-wild audio are used for evaluation.
As shown in Figure \ref{fig:fmfl_vis}, TDM may produce unnatural facial expressions, misaligned lip movements, and visible artifacts such as excessive lip closure or excessive mouth opening, all of which significantly degrade the perceptual quality. In contrast, fine-tuning using FMFL can effectively mitigate these issues, resulting in more natural and perceptually pleasing 3D facial animations.
Notably, FMFL fine-tuning guided by FMReward neither directly penalizes large facial and head movements nor biases the generation toward static motions; instead, it suppresses unnatural or excessive motions to produce movements that better align with human perceptual preferences.

\noindent\textbf{User Study.}
To comprehensively evaluate the performance of our model, we further conduct a user study comparing it with state-of-the-art methods and our unfine-tuned TDM model. Specifically, we randomly select 13 audio clips from the TFHP test set and use them to animate all methods for comparison. Participants are asked to choose the better one in each pairwise comparison of facial motions generated by our method and a baseline method, and the final preference for each pair is determined by majority voting. A total of eleven participants take part in the user study. As shown in Figure \ref{fig:fmfl_user_study}, our model outperforms all baseline methods in the majority of samples, demonstrating the effectiveness of fine-tuning with the FMFL algorithm and the superior performance of our final model.

\section{Conclusion}
In this paper, we explore the problem of human preference alignment in audio-driven 3D facial animation for the first time.
To begin with, we construct FMPair, the first human preference dataset for audio-driven 3D facial animation, comprising 65,574 annotated 3D facial motion pairs from 8,834 audio clips.
Along with FMPair, we propose FMReward, an automatic perceptual model designed to predict quality score from audio and 3D facial motion.
Based on FMReward, we further introduce FMFL, a direct fine-tuning algorithm to optimize diffusion-based audio-driven 3D facial animation models for better alignment with human preferences.
Extensive experiments demonstrate the superiority of FMReward in aligning with human preferences and the effectiveness of FMFL in improving the perceptual quality of audio-driven 3D facial animation.
We hope that the human preference alignment concept will pave the way for more natural and perceptually faithful audio-driven 3D facial animation.

\bibliographystyle{IEEEtran}
\bibliography{refs}

\vfill

\end{document}